\documentclass{article}
\usepackage{longtable}
\usepackage[final]{neurips}
\usepackage[utf8]{inputenc} 
\usepackage[T1]{fontenc}    
\usepackage{hyperref}       
\usepackage{url}            
\usepackage{booktabs}       
\usepackage{amsfonts, amsmath}       
\usepackage{nicefrac}       
\usepackage{microtype}      
\usepackage{lipsum}
\usepackage{graphicx}
\usepackage{enumitem}
\usepackage{color}
\usepackage{verbatimbox}
\usepackage{listings}
\usepackage{xcolor}
\usepackage[numbers]{natbib}
\usepackage{xspace}
\usepackage{tabularx}
\usepackage{wrapfig}
\usepackage{arydshln}

\definecolor{codegreen}{rgb}{0,0.6,0}
\definecolor{codegray}{rgb}{0.5,0.5,0.5}
\definecolor{codepurple}{rgb}{0.58,0,0.82}
\definecolor{backcolour}{rgb}{0.95,0.95,0.95}

\setlist[itemize]{leftmargin=*}
\setlist[enumerate]{leftmargin=*}
\graphicspath{ {./images/} }

\lstdefinestyle{mystyle}{
  backgroundcolor=\color{backcolour}, commentstyle=\color{codegreen},
  keywordstyle=\color{magenta},
  numberstyle=\tiny\color{codegray},
  stringstyle=\color{codepurple},
  basicstyle=\ttfamily\footnotesize,
  breakatwhitespace=false,         
  breaklines=true,                 
  captionpos=b,                    
  keepspaces=true,                 
  numbers=left,                    
  numbersep=5pt,                  
  showspaces=false,                
  showstringspaces=false,
  showtabs=false,                  
  tabsize=2
}

\usepackage[strict]{changepage}
\usepackage{xcolor}
\usepackage{framed}
\definecolor{demonstrationshade}{rgb}{0.95,0.95,0.95}
\definecolor{promptshade}{rgb}{0.95,0.95,0.95}

\usepackage{todonotes}
\usepackage{xcolor}

\newcommand{\method}{ContextHub\xspace}
\newcommand{\llms}{LLMs\xspace}
\newcommand{\dyval}{DyVal\xspace}

\title{Disentangling Logic: The Role of Context in Large Language Model Reasoning Capabilities}

\author{
 Wenyue Hua* \\
 Rutgers University
 \And
 Kaijie Zhu* \\
 Microsoft
 \And
 Lingyao Li \\
 University of Michigan
 \And
 Lizhou Fan \\
 University of Michigan
 \And
 Shuhang Lin \\
 Rutgers University
  \And
 Mingyu Jin \\
 Rutgers University
 \And
 \hspace{-10pt}Haochen Xue \\
 \hspace{-10pt}The University of Liverpool
 \And
 \hspace{-10pt}Zelong Li \\
 \hspace{-5pt}Rutgers University
 \And
 Jindong Wang \\
 Microsoft
 \And
 Yongfeng Zhang\thanks{Corresponding Emails: wenyue.hua@rutgers.edu, kaijiezhu11@gmail.com, yongfeng.zhang@rutgers.edu} \\
 Rutgers University
}

\begin{document}

\maketitle

\begin{abstract}
This study intends to systematically disentangle pure logic reasoning and text understanding by investigating the contrast across abstract and contextualized logical problems from a comprehensive set of domains. We explore whether \llms demonstrate genuine reasoning capabilities across various domains when the underlying logical structure remains constant. We focus on two main questions (1) Can abstract logical problems alone accurately benchmark an LLM's reasoning ability in real-world scenarios, disentangled from contextual support in practical settings? (2) Does fine-tuning \llms on abstract logic problem generalize to contextualized logic problems and vice versa? To investigate these questions, we focus on standard propositional logic, specifically propositional deductive and abductive logic reasoning. In particular, we construct instantiated datasets for deductive and abductive reasoning with 4 levels of difficulty, encompassing 12 distinct categories or domains based on the categorization of Wikipedia. Our experiments aim to provide insights into disentangling context in logical reasoning and the true reasoning capabilities of \llms and their generalization potential. The code and dataset are available at: \url{https://github.com/agiresearch/ContextHub}.
\end{abstract}

\section{Introduction}
Large language models (\llms) \cite{brown2020language, chowdhery2023palm, chung2022scaling} have demonstrated significant potential in reasoning capabilities across a variety of reasoning benchmarks~\cite{cobbe2021training, hendrycks2021measuring, wei2022emergent, liang2022holistic, srivastava2023beyond, zhu2023dyval, fan2023nphardeval, fu2024isobench}, broadening their potential applications in fields such as psychology, education, and social sciences~\cite{gandhi2024understanding, li2024hot, fan2023datachat}. The widespread use of \llms accentuates the necessity of rigorously evaluating their reasoning abilities, particularly in context-rich scenarios that mirror real-world complexities.

While assessments on abstract logical problems~\cite{sawada2023arb, zhu2023dyval} showcase \llms' theoretical reasoning capacities, they do not entirely capture their practical utility in real-life applications where context drastically affects outcomes. Conversely, focusing exclusively on context-specific tasks may conceal the fundamental mechanisms that empower \llms to process and reason with information. Thus, exploring the balance between contextualized and abstract reasoning is vital for responsibly advancing LLM technology and ensuring its effectiveness across various domains.

To this end, we introduce \method—a pioneering benchmark designed to meticulously disentangle and evaluate the core reasoning capabilities of \llms from the influences of contextual information. By leveraging a dual-assessment framework, \method compares \llms' performance on identical logical constructs within both abstract and richly contextualized settings. This approach not only highlights the differential impacts of context on reasoning but also provides a scalable and flexible methodology that can be adapted across various domains and LLM architectures.
Our approach aims to address two main questions:
\begin{enumerate}
    \item \textit{Evaluation disentanglement: how accurate and robust is it to evaluate \llms' reasoning abilities using abstract logic problems or various contextualized logic problems?} By comparing the performance of \llms on abstract and contextualized logical problems, we can gain a better understanding of the role of context in \llms' reasoning abilities.
    \item \textit{Fine-tuning disentanglement: how does model generalization differ when fine-tuning \llms using abstract logic problems or contextualized logic problems?} By comparing the performance of \llms on unseen abstract and contextualized logic problems, we can gain insights into the types of data that are most effective for improving \llms' reasoning abilities while maintaining consistent performance across different domains.
\end{enumerate}

We first utilize DyVal\cite{zhu2023dyval} to generate $4$ different difficulty levels of formal logic templates. Then, we use advanced \llms to contextualize these logical templates.  
Our key finds are:
\begin{enumerate}
    \item The relative performance of \llms on abstract logic and corresponding instantiated logic is dependent on model size or general model performance. Stronger models tend to perform better on abstract logic, while smaller models typically rely on contextual cues.
    \item The domain of contextualization has a statistically significant impact on model performance. This suggests that the choice of contextualization domain can affect the accuracy and reliability of \llms for logical reasoning tasks.
    \item The generalization power of abstract logic data is limited compared with that of instantiated logic data. This indicates that \llms fine-tuned on instantiated logic data may be better equipped to handle a wider range of logical reasoning tasks, including those that involve real-world scenarios and contextual cues.
\end{enumerate}

To sum up, this paper makes the following contributions:
\begin{enumerate}
    \item \textbf{Flexible instantiation evaluation framework.} \method enables researchers to easily create contextualized logical datasets for their own studies, allowing for more comprehensive and diverse evaluations of \llms' reasoning abilities. 
    \item \textbf{Extensive benchmarking and analysis.}  Empirical evidence demonstrates that abstract logic assessments do not fully reflect an \llms' reasoning capabilities and different instantiations impose significant effect on model performance.
    \item \textbf{Investigation into data generalization potential.} Extensive experiments show that abstract logic has limited generalization power while contextually instantiated show better generalization ability.
\end{enumerate}


\section{Related Work}
Efforts for evaluating \llms reasoning abilities have been intensified significantly across numerous disciplines, including biomedical informatics \cite{lievin2024can, chen2024evaluating, jin2024health}, humanities \cite{hua2023war, lin2024battleagent, jin2024if}, and social sciences \cite{ziems2024can, gandhi2024understanding, li2024hot, fan2023datachat}.
Numerous studies have instantiated the problem of logical reasoning based on reasoning-dependent datasets, such as deduction, induction or abduction, and studies solving the tasks with neural models \cite{pan2023logic, li2024reason, dasgupta2022language}. 
Despite the promising performance that LLMs have shown on certain reasoning tasks and those techniques that can help improve LLMs' reasoning abilities, it remains unclear whether LLMs have generalizable logical reasoning abilities and to what extent they are capable of logical reasoning~\cite{tang2023paradox}. In this regard, Valmeekam et al.~\cite{valmeekam2022large} stated that LLMs were still fell short of delivering satisfactory performance in common planning and reasoning tasks that are typically straightforward for humans to perform. This limitation was also highlighted by Wei et al.~\cite{wei2022chain} that although CoT can stimulate the thought processes of human reasoners, it does not answer whether the neural network is actually reasoning. 

Other studies have also highlighted the limitations of modern LLMs in performing logical reasoning tasks. For example, Tang et al.~\cite{tang2023paradox} found that LLaMA2, relied on template matching to respond to reasoning queries but failed to generalize to novel logic rules, as demonstrated through experiments on Symbolic Trees and ProofWriter. This led them to question whether modern LLMs have truly mastered inductive, deductive, and abductive reasoning abilities akin to human intelligence. In a recent benchmark study, Saparov and He~\cite{saparov2022language} presented a synthetic question-answering dataset called PrOntoQA to explore the logical reasoning ability of LLMs. Their analysis revealed that while LLMs were capable of making correct individual deduction steps, they encountered difficulties in exploring the solution space when presented with multiple valid deduction steps.

Based on our review of current studies, the prevalent focus on question answering and mathematical problems in current benchmarks may not sufficiently capture the essence of reasoning - the ability to logically process and deduce information beyond memorized knowledge or instantiated cues. In particular, two specific questions regarding the logical reasoning abilities of LLMs remain unclear. First, it is unclear whether LLMs genuinely comprehend logical reasoning patterns within the text and generate answers through reasoning, or if they merely follow the textual instructions and answer based on prior knowledge. Second, it is uncertain whether current reasoning benchmarks accurately assess the true reasoning abilities exhibited by LLMs in practical scenarios. To investigate these questions, we focus on propositional logic, specifically deductive and abductive logic reasoning in this study.

\begin{figure}
    \centering
    \includegraphics[scale=0.3]{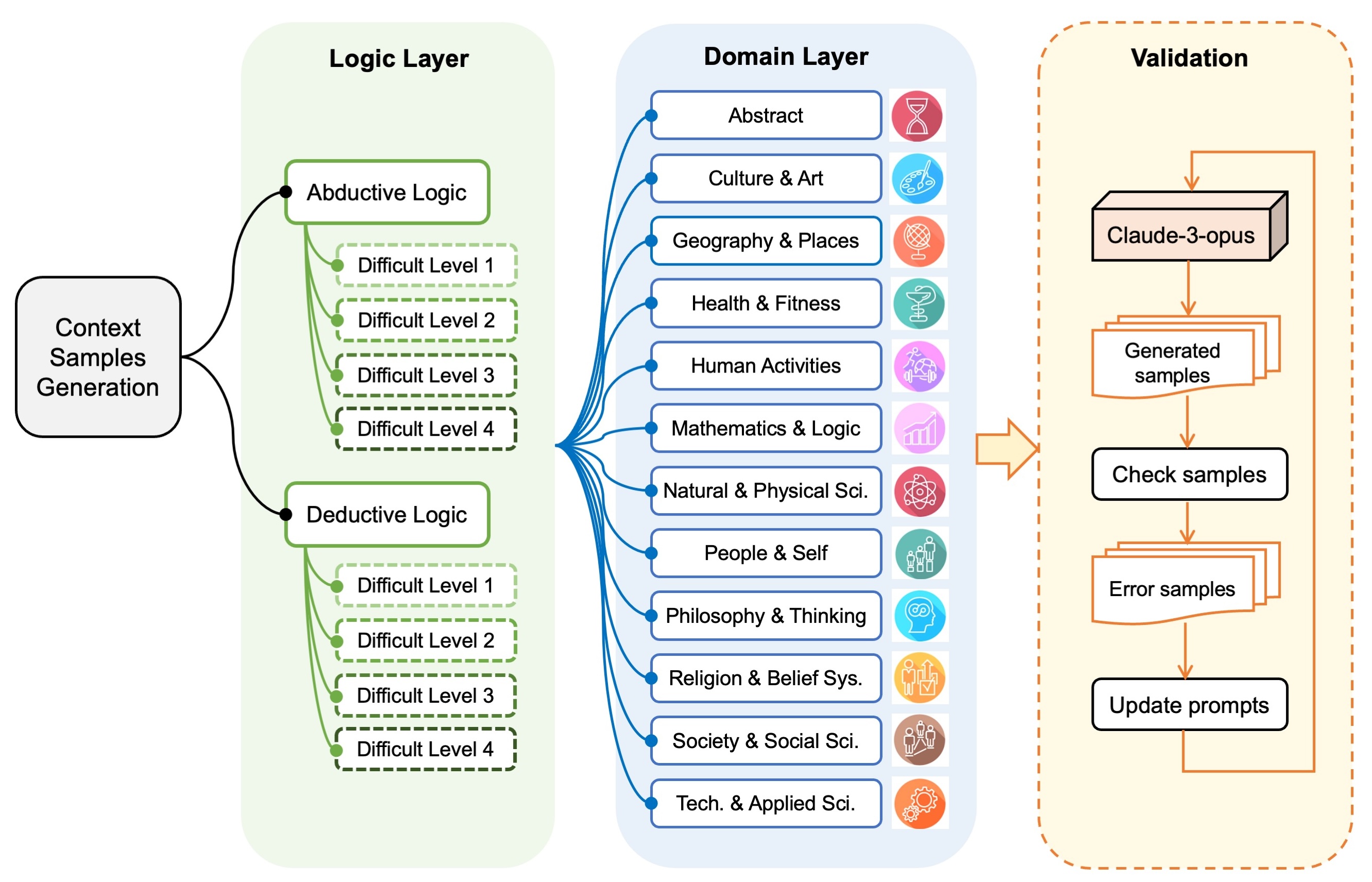}
    \caption{Benchmark Construction Procedure}
    \label{fig:construction}
\end{figure}

\section{Benchmark Construction}
In this paper, we focus on two types of propositional logical reasoning tasks: deductive logic and abductive logic. Deductive logic is a method of reasoning from one or more general statements (premises) to reach a logically certain conclusion, where abductive logic involves starting with an observation and then abduce the truth value of premises.

As illustrated in Figure \ref{fig:construction}, constructing instantiated logical reasoning benchmarks involves three steps:
\begin{enumerate}
    \item \textbf{Creating Formal Logical Reasoning Question Templates.} We first construct abstract deductive and abductive logic questions spanning $4$ difficulty levels based on \dyval~\cite{zhu2023dyval}. \dyval utilize graph structure to dynamically construct logic questions. We call these questions formal logic template for further contextualization.
    \item \textbf{Instantiation.} Each logical template $T$ will then be instantiated in 12 different domains. In each domain, we randomly select one of its sub-categories to diversify the context. We then ask \llms to instantiate the logical template $T$ using the context in each sub-category. 
    \item \textbf{Quality Control.} To ensure the correctness of the instantiated evaluation samples, we implement a two-step quality control process. Initially, the samples are assessed by an advanced LLM, \textit{Claude 3 Opus}, which validates the samples against specified constraints. Subsequently, for each level of difficulty, 220 randomly selected samples undergo a further review by $5$ human experts.

\end{enumerate}

\subsection{Creating Formal Logical Reasoning Question Templates}
We generate formal logic templates $\mathcal{X}$ is based on the dynamic evaluation framework : \dyval~\cite{zhu2023dyval}. For deductive and abductive logic, \dyval utilize tree structure to to generate template samples on the fly with controllable difficulty. The tree structure naturally align with the inference process of a logic reasoning question. Take deductive logic as an example, the premises are given by the leaf nodes, where the intermediate nodes represents the intermediate inference steps, the final result is shown by the root node.

Tree-based \dyval consists of three components:
\begin{enumerate}
    \item Constraint $\mathcal{C}$. It aims to modulate the evaluation samples' complexity and validity. In our experiment, we define the complexity level of a logical question by its depth of the generated tree. The validity constraints ensure the correctness of the generated logic question, for example, constrain the `NOT' operation to have only one children node;
    \item Tree generation algorithm $\mathcal{G}$. After defining the constraints, the generation algorithm $\mathcal{G}$ generates fixed complexity evaluation samples following the constraint $\mathcal{C}$, during the generation process, the final answer is also be calculated;
    \item Description function $\mathcal{F}$. It translate each node in the graph into natural languages and finally form all nodes to a logical reasoning question. For example, in deductive logic, for a leaf node `A' with truth value `True', it will be translated as ``A is True.'', for a non-leaf node `C' with `OR' operation and its children `A' and `B', it will be translated as ``(A OR B) - C'', where `-' means deductive operation.
\end{enumerate}

In our experiments, we defined $4$ levels of difficulty with depth equals to $(2, 3, 4, 5)$, respectively. The width are set to be $2$. For level 1 difficulty, there can only be 10 different deductive questions and 6 abductive questions. For other levels, we generate 40 deductive logic templates and 40 abductive logic templates. To balance data distribution, we generate the same number of datapoints assigned with the truth values of True, False, and N/A respectively. 

\subsection{Contextual Instantiation}

\paragraph{Domains of contextualization} We instantiate the above formal logic templates in the below contextual domains following the categorization of Wikipedia~\cite{Wiki}:
\begin{lstlisting}[language=HTML, caption=Categories of Wikipedia]
Culture and the arts, Geography and places, Health and fitness, Human activities, Mathematics and logic, Natural and physical science, People and self, Philosophy and thinking, Religion and belief systems, Society and social sciences, Technology and applied sciences. 
\end{lstlisting}
The domain of ``History and events'' is removed because instantiated sentences are often about known fact and the question can be answer without going through the reasoning process, thus nullify the reasoning problem.

Each instantiation of a domain is created based on a randomly selected sub-categories in the domain from above based on sub-categories established in Wikipedia to encourage diversity and specification. For example, ``Culture and the arts'' has the following sub-categories:
\begin{lstlisting}[language=HTML, caption=Sub-categories of Culture and the arts in Wikipedia]
Classics, Critical theory, Cultural anthropology, Clothing, Folklore, Food and drink culture, Language, Literature, Museology, Mythology, Philosophy, Popular culture, Science and culture, Traditions, Arts and crafts, Celebrity, Censorship in the arts, Festivals, Humor, Literature, Museums, Parties, Poetry, Circuses, Dance, Film, Music, Opera, Storytelling, Theatre, Architecture, Comics, Crafts, Design, Drawing, Film Animation, New media art, Painting, Photography, Sculpture, Board games, Card games, Dolls, Puppetry, Puzzles, Role-playing games, Video games, Air sports, American football, Association football, Auto racing, Baseball, Basketball, Boating, Boxing, Canoeing, Cricket, Cycling, Exercise, Fishing, Golf, Gymnastics, Hobbies, Horse racing, Ice hockey, Lacrosse, Olympic Games, Rugby league, Rugby union, Sailing, Skiing, Swimming, Tennis, Track and field, Walking trails, Water sports, Whitewater sports
\end{lstlisting}

\paragraph{Contextualization process} 
After obtaining the formal logic templates, for each domain, we first randomly selected one sub-category $c$, then we ask \llms (in our experiment, Claude-3-Opus) to instantiate each variable in the original logic templates with the relevant context in the selected sub-category. This contextualizatin process is divided into 2 steps:
\begin{enumerate}
    \item Variable-based Transformation: $\mathcal{T}_v$. For each variable $\mathcal{V}$ contained in the logic template $\mathcal{X}$, a instantiated sentence $s_\mathcal{V}$ is generated by $\mathcal{T}_v(c, \mathcal{V}\in\mathcal{X})$. For example, a leaf node variable $\mathcal{V}$ can be instantiated as ``Alice studied hard for the following math test'' in the sub-category of ``Mathematics Education'' in the category of ``Mathematics and Logic''.
    \item Template-based Transformation: $\mathcal{T}_t$. After generating $\{s_\mathcal{V}\}$ for all $\{\mathcal{V}\}$ in the template $\mathcal{X}$, a coherent natural language description will be generated by $\mathcal{T}_t(\{s_\mathcal{V}\}, \mathcal{X})$ by forming a instantiated version of the original formal logic template.
\end{enumerate}

\subsection{Abstract Instantiation}
Other than the 11 contextual domains from Wikipedia, we also create an ``abstract'' domain where we simply substitute by heuristic rules the propositional variables in the formal logic template with arbitrary character sequences of varying lengths, ranging from 3 to 5. The purpose of creating this domain is to augment the number of datapoints expressed in an abstract form, thereby enabling a fair comparison with other contextualized domains in terms of sample size. Furthermore, by employing multiple instantiations, we can mitigate the impact of any potential outliers and obtain a more reliable and generalizable estimate of the performance of abstract data as we have only 256 formal logic templates in total.

Below is an example of instantiation in Table \ref{tab:example} of a formal logic template of difficulty level 1, where propositional variables are represented by strings such as "aaa", "aab", and "aac".
\begin{lstlisting}
(aaa or aab) - aac.  Given aac is False, what is the value of aab?
\end{lstlisting}

We provide an abstract instance and a contextualized instance on the domain of ``Geography and Places'', where we provide the instantiations of each proposition in the template and the final combined logic reasoning task based on propositional instantiations. More examples can be found in Appendix \ref{app:data_examples}.
\begin{table}[!ht]
    \centering
    \begin{tabular}{p{0.25\textwidth}p{0.68\textwidth}}
    \toprule
    Level 1 - Abstract &  Level 1 - Geography and Places \\
    \midrule
    aaa: vxkgr  & aaa: The terrain has experienced significant uplift.  \\
    aab: caunc  & aab: Powerful erosional forces have shaped the land.  \\
    aac: ybyz  & aac: The area features tall, steep mountains.  \\
    reasoning task: (vxkgr or caunc) $\rightarrow$ ybyz. Given ybyz is False, what is the value of caunc? & reasoning task: If an area of land has experienced significant uplift or been shaped by powerful erosional forces, then the terrain will feature tall, steep mountains. Given that the area does not have tall, steep mountains, can it be determined if powerful erosional forces have shaped the land?  \\
    \midrule
    \end{tabular}
    \caption{Example of abstract and instantiated logic reasoning task based on the original formal logic template.}
    \label{tab:example}
\end{table}

\subsection{Data Statistics}
Consequently, for each formal logic template, there are a total of 12 domains used for contextualization, including the abstract domain.
For each domain, we generate 5 distinct instantiations. Thus, for level 1 abductive logic, we have in total 360 datapoints; for level 2 deductive logic, we have in total 600 datapoints. For other levels of abductive and deductive logic, we have 2,880 datapoints. Thus the whole benchmark contains 18,240 datapoints.


\subsection{Quality Control}
The quality verification of our instantiated benchmarks is managed using a hybrid model involving Claude-3 Opus, and a diverse panel of $5$ human annotators. These verification steps are implemented to maintain a high standard of quality and relevance in our benchmarks, ensuring that they not only test logical reasoning but also engage with the domain knowledge in a meaningful way. Notice that since the "abstract" instantiations are created simply by rule-based heuristics, no further verification is needed and thus quality control is only conducted in contextually instantiated reasoning questions.

\paragraph{\llms verification}
Ensuring the validity of our benchmarks involves three primary checks, which are critical to establishing the reliability of the logic problems within the instantiated settings. These checks are designed to assess the problems for fundamental logical soundness and adherence to rational thinking patterns:

\begin{itemize}
    \item \textbf{Common Sense Checking:} This verification step assesses whether the contextually instantiated logic problem relies on universally recognized knowledge, such as common sense facts that do not require logical reasoning to solve. The purpose of this check is to ensure that the questions demand genuine logical inference rather than mere recognition of widely known facts. 

    \item \textbf{Sensibility Checking:} Each logic problem is scrutinized for its sensibility and coherence. This process ensures that the scenarios and the associated questions are coherently constructed and present a clear, understandable challenge to models without any internal contradictions or ambiguous phrasing.

    \item \textbf{Tautology Checking:} This check is crucial for identifying any statements within the logic problems that are inherently tautological. A tautology in this context would be a condition or a statement that is true in every possible interpretation, thus rendering the problem trivial or meaningless. This step helps in maintaining the intellectual rigor and challenge expected from a benchmark in logical reasoning.
\end{itemize}

\paragraph{Human verification}
In addition to automated checks, human verification plays a pivotal role in ensuring the quality and applicability of our instantiated benchmarks. For detailed demographics of these trained annotators, refer to Appendix \ref{app:verification}, which includes their degrees, gender, age, and other relevant information. The human verification focuses on two critical aspects:

\begin{enumerate}
    \item \textbf{Template Adherence:} The first aspect of human verification assesses whether the instantiated logic accurately follows the structure and intent of the original formal logic template. This check is essential to ensure that the fundamental logical framework remains intact and that the contextual adaptations do not alter the core logical challenges intended by the original template. Annotators evaluate the fidelity of the instantiated problems to their original versions, confirming that the essential logical elements are preserved.

    \item \textbf{Fact Reckoning:} The second aspect evaluates whether the contextually instantiated template accounts for commonly known facts, which could potentially simplify the logical reasoning required to answer the question. This step is crucial to ensure that the problems require genuine logical deduction rather than mere recall or memorization. It seeks to eliminate any instances where the answer to a problem might be directly inferred from general knowledge or trivial facts, thereby maintaining the complexity and educational value of the benchmark.
\end{enumerate}

The dual-layer verification process ensures that our benchmarks not only test logical reasoning skills but also do so in a manner that is both challenging and fair.

\section{Experimental Design}
The experiments in this study are divided into two parts: benchmarking and fine-tuning. The benchmarking part aims to evaluate model performance across different domains and also compare the performance of contextualized logic to that of abstract logic, examining whether LLM truly understands the underlying logic structure regardless of the contexts or instantiations. The data points generated for instantiated logic can also be used for fine-tuning models, which is the focus of the fine-tuning part. This part explores various aspects of fine-tuning using synthetic data, including the use of abstract logic instantiations or contextualized logic instantiations, model scaling on the effect of generalization, and the domains of contextualized data.

We benchmark with three distinct series of well-trained models: Qwen-1.5 \cite{qwen}, LLaMA-2 \cite{touvron2023llama}, and Yi-1.5 \cite{young2024yi}. The selection of these models is primarily based on their varying sizes, which enables us to conduct a comprehensive analysis of the impact of model size and scaling on the performance of LLMs. The Qwen-1.5 series includes the following models: Qwen1.5-0.5b, Qwen1.5-1.8b, Qwen1.5-4b, Qwen1.5-7b, Qwen1.5-14b, Qwen1.5-32b, Qwen1.5-72b, and Qwen1.5-110b. The LLaMA-2 series includes the following models: LLaMA2-7b, LLaMA2-13b, and LLaMA2-70b. The Yi-1.5 series includes the following models: Yi1.5-6b, Yi1.5-9b, and Yi1.5-34b.

In the fine-tuning stage of our study, we employ four models: Qwen1.5-0.5b, Qwen1.5-7b, Qwen1.5-14b, and GPT-3.5-turbo. Our goal is to investigate the factors that impact model generalization for logic reasoning, and to this end, we explore several fine-tuning settings: (1) fine-tune the models solely on abstract data, (2) fine-tune the models on sampled contextualized data sampled from all domains, and (3) fine-tune the models on contextualized data from single-domains. Results from (1) and (2) allow us to investigate the generalization ability of abstract data and to compare it to the generalization ability of contextualized data; while results from (3) allow us to investigate the impact of domain specificity and domain diversity on generalization. 

\paragraph{Evaluation Metrics for Benchmarking}
\label{sec:eval}

To assess the reasoning capabilities of LLMs, we employ the average F1 score. The calculation of the average F1 score involves determining the average of the F1 scores for data points ($d$) that possess the same truth values. For datapoints with identical ground truth ($gt$) truth value $\mathcal{T}$, the F1 score is computed by first ascertaining the true positive ($T_p^{\mathcal{T}}$), false positive($F_p^{\mathcal{T}}$), and false negative($F_n^{\mathcal{T}}$):

\begin{align}
    T_p^{\mathcal{T}} &= {\{d\in D | f(d) = gt(d), gt(d) = \mathcal{T}\}} \\
    F_p^{\mathcal{T}} &= {\{d\in D | f(d) \neq gt(d), f(d) = \mathcal{T}\}}\\
    F_n^{\mathcal{T}} &= {\{d\in D | f(d) \neq gt(d), gt(d) = \mathcal{T}\}}
\end{align}

F1$^{\mathcal{T}}$ for for datapoints with the truth value $\mathcal{T}$ is then computed by:

\begin{align}
    F1^{\mathcal{T}} = \frac{2T_p^{\mathcal{T}}}{2T_p^{\mathcal{T}} + F_p^{\mathcal{T}} + F_n^{\mathcal{T}}}
\end{align}

The average F1 score for the entire dataset is calculated by determining the average of the F1$^{\mathcal{T}}$ scores for all possible truth values $\mathcal{T}$. 

\paragraph{Hyperparameters for Finetuning}
We leverage QLora \cite{dettmers2024qlora} for finetuning on open-source models. Other relevant hyperparameters are: epochs = 3, warmup proportion = 0.01, learning rage = 3e-4, weight decay = 0.01, lora rank = 64, lora dropout = 0.05, lora alpha = 16, batch size = 4, accumulate gradient steps = 8. 

\begin{figure}
    \centering
    \includegraphics[scale=0.6]{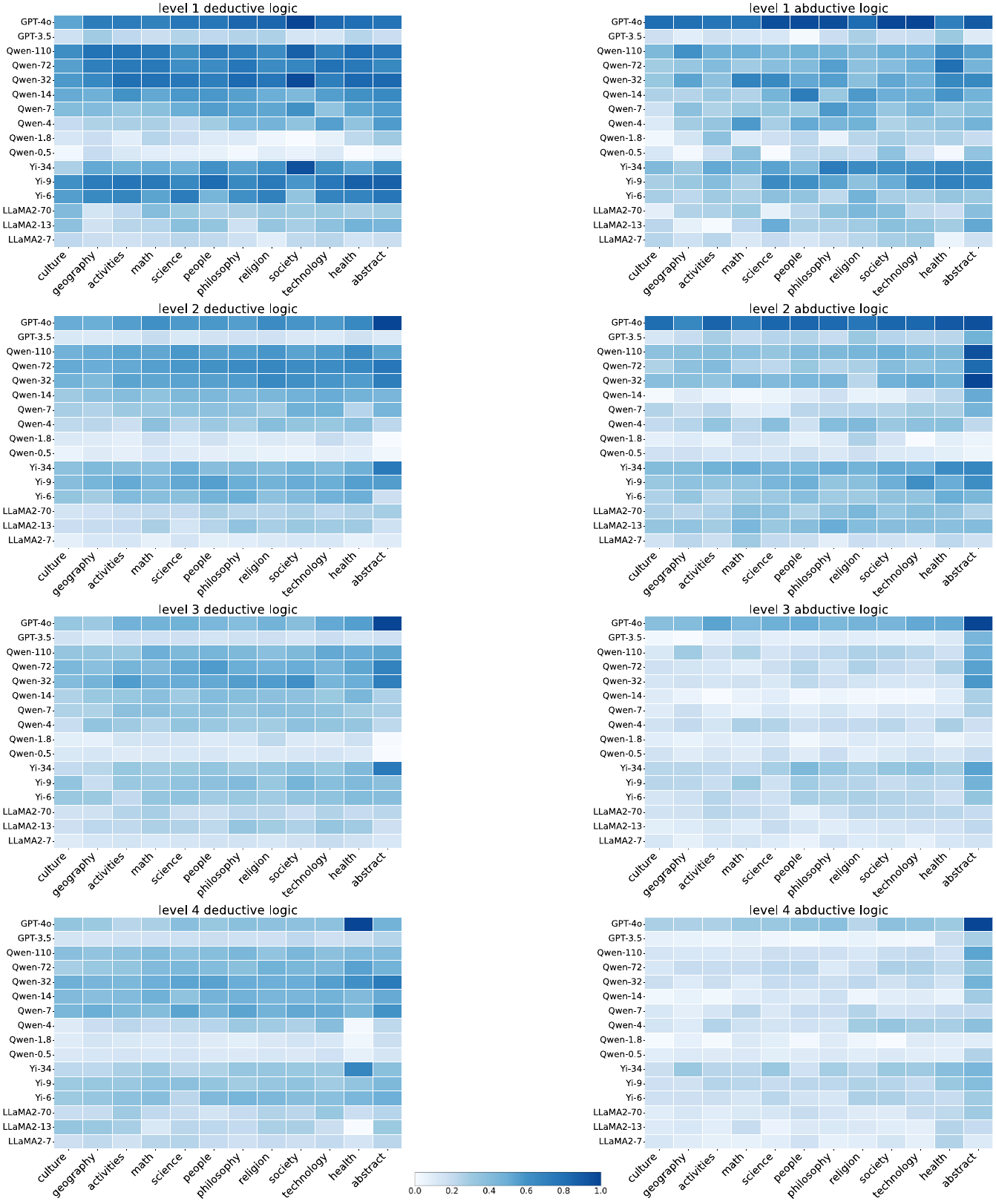}
    \caption{Main Benchmark Performance}
    \label{fig:general_performance}
\end{figure}

\section{Experiment Analysis}

In this section, we present a comprehensive analysis of the results obtained from our experiments. Specifically, we focus on two main areas: benchmark results and fine-tuning results. The benchmark results provide a general analysis of the performance trends, as well as a statistical analysis of the impact of domain on model performance. Meanwhile, the fine-tuning results offer insights into the factors that influence model generalization for logic reasoning. By examining these two areas in detail, we aim to provide a thorough understanding of the behavior of large language models in the context of logic reasoning.

\subsection{Benchmarking}

\paragraph{Overview of Model Performance}
The evaluation results are presented in \figurename~\ref{fig:general_performance}. The performance across models varies, as depicted in the heatmap distribution. At a granular level, GPT-4o models frequently appear to excel, particularly in higher difficulty levels. In contrast, smaller models like Qwen-0.5 and Qwen-1.8 often lag, struggling notably with abstract reasoning tasks. This disparity underscores the influence of model size. When aggregating results across all models and logic levels, certain domains consistently present more challenges. Specifically, the domains of Math and Philosophy appear to be the most demanding, likely due to their intrinsic requirement for deep logical structuring and abstract reasoning. Conversely, the domain labeled People generally shows the best performance, suggesting that the models are better attuned to reasoning about human-centered contexts, which might be less abstract or feature more contextual cues. The performance difference among domains are tested to be statistically significant as discussed below.

\begin{figure}[!ht]
    \centering
    \includegraphics[scale=0.35]{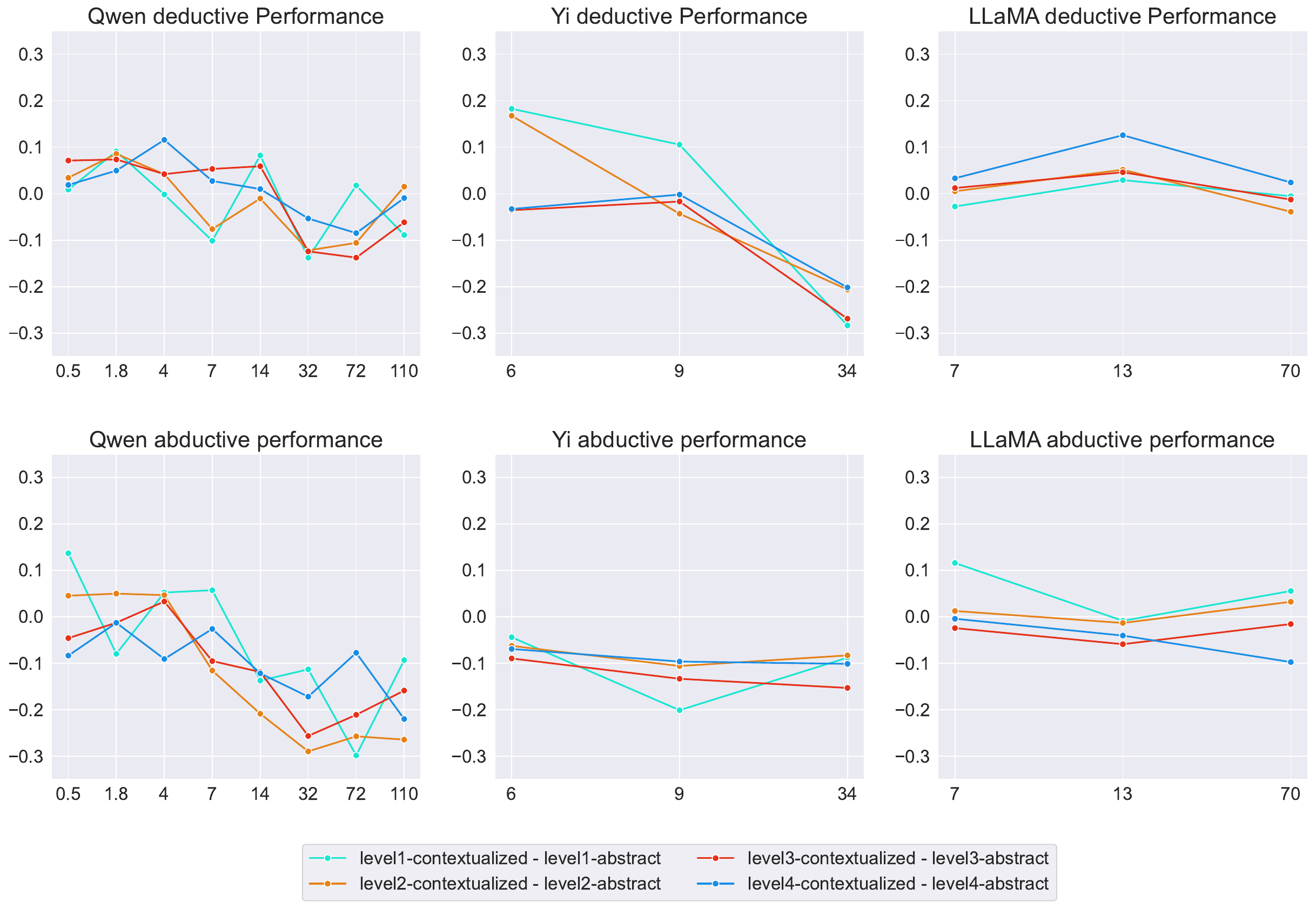}
    \caption{Abstract performance vs. instantiated performance}
    \label{fig:abstract_vs_instantiated}
\end{figure}

\paragraph{Influence of Model Size}
A pivotal observation from our data is the interaction between model size and sample type, presented in Figure \ref{fig:abstract_vs_instantiated}. Larger models demonstrate a marked proficiency in abstract logical reasoning samples compared to their performance with their corresponding instantiated samples. This trend holds regardless of the difficulty level, suggesting that as models scale, their ability to decipher and apply abstract logic patterns improves significantly. While smaller models, like Qwen-0.5, Yi-6, LLaMA-7, demonstrate either better performance on instantiated samples than abstract samples or less difference between these two types. This discovery differs from previous observations \cite{tang2023paradox, saparov2022language} which in general state that LLMs are better at instantiated data.

\paragraph{Inter-domain Disparities}
Further analysis of specific model performance within different domains reveals notable patterns. For abstract reasoning tasks, performance is highly variable: smaller models like Qwen-0.5 and Qwen-1.8 perform significantly worse, while larger configurations often excel. In the domain of Math, both Yi and Qwen series models exhibit consistently lower performance, reinforcing the notion of this domain's complexity. Interestingly, there is a general trend from observation where models that generally perform well show more pronounced disparities across domains, suggesting that higher capabilities amplify domain-specific challenges or advantages.




\begin{figure}
    \centering
    \includegraphics[scale=0.3]{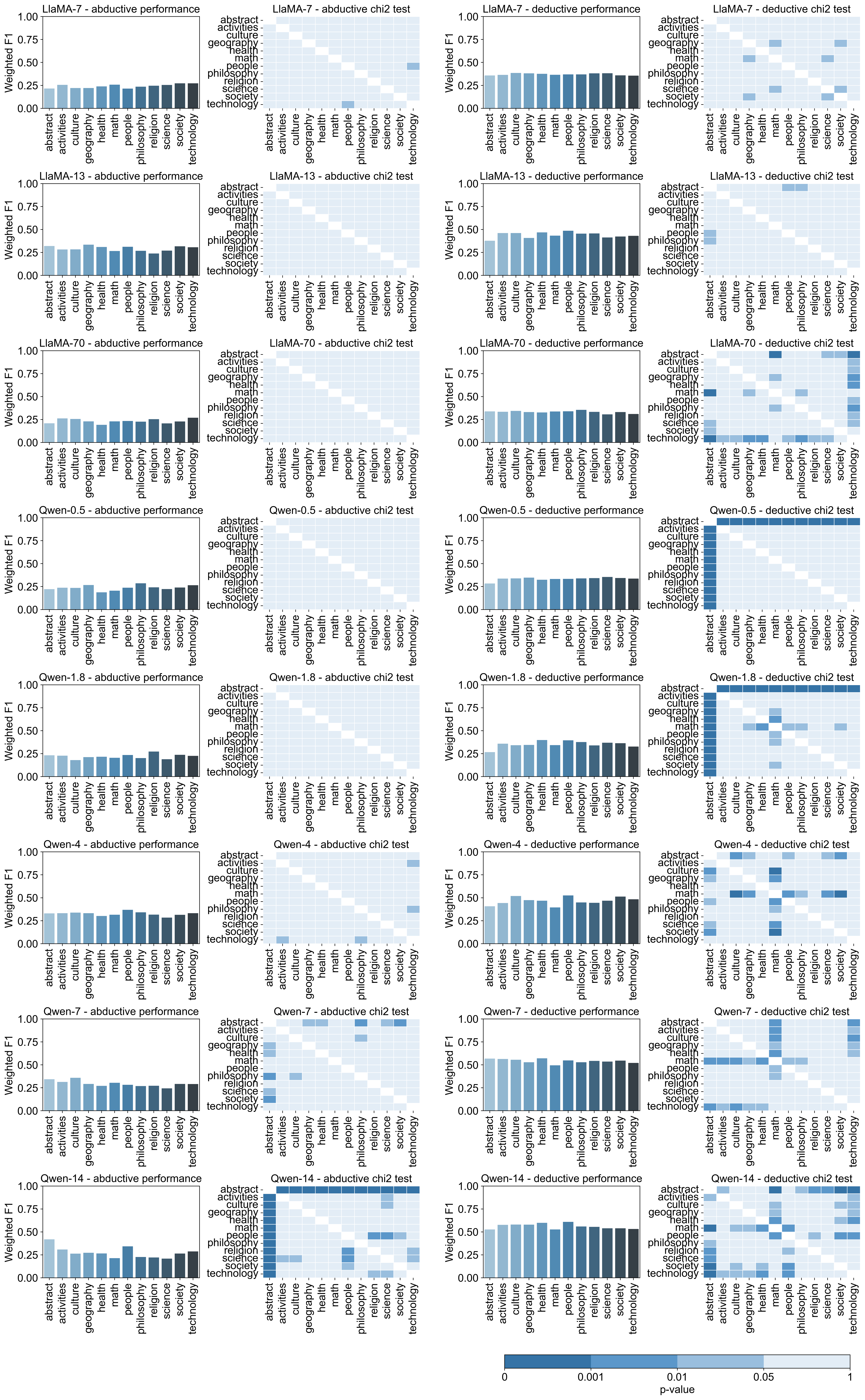}
    \caption{Results of weighted F1-score and Chi-square test}
    \label{fig:statistics_1}
\end{figure}

\begin{figure}
    \centering
    \includegraphics[scale=0.3]{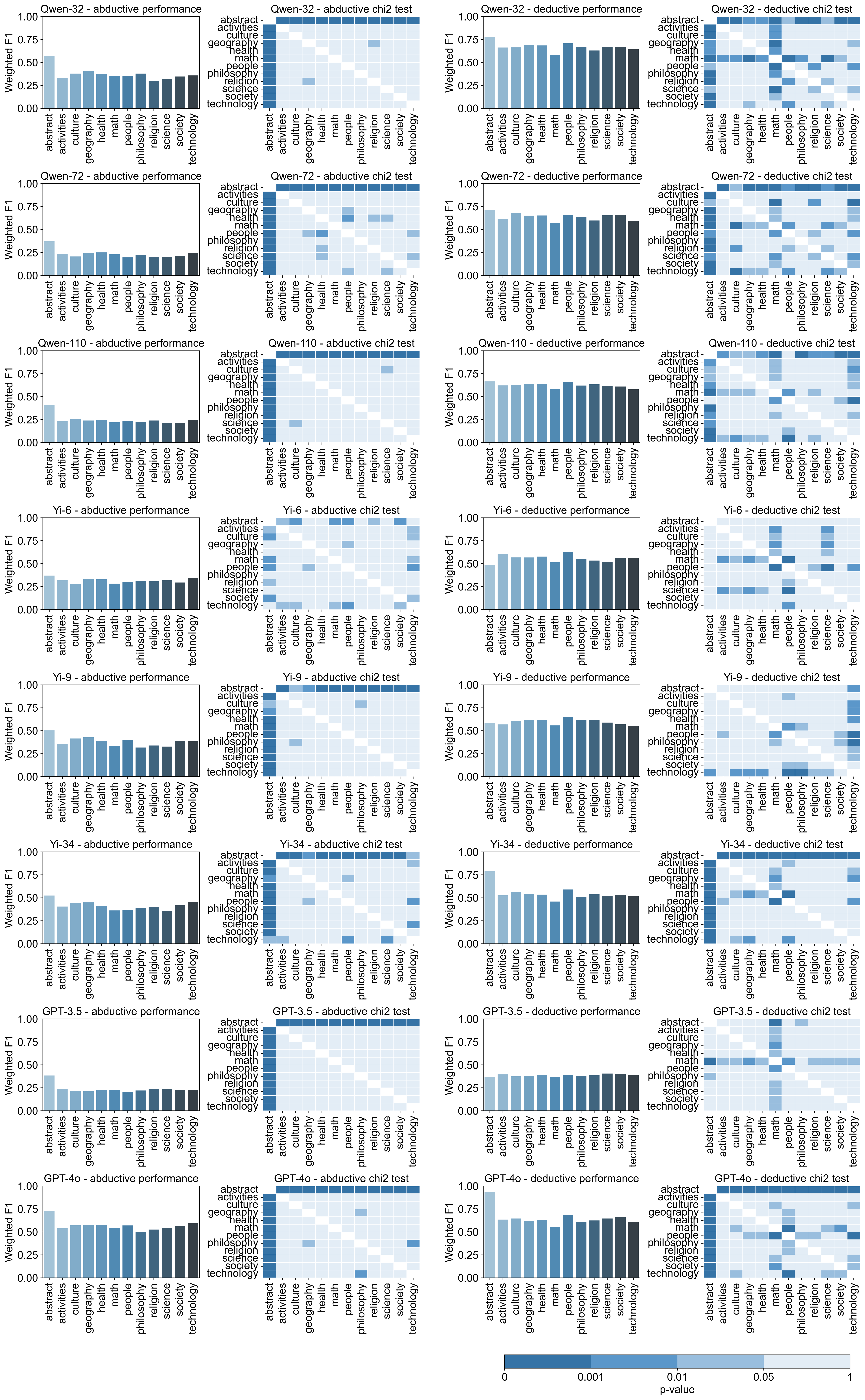}
    \caption{Results of weighted F1-score and Chi-square test (Cont.)}
    \label{fig:statistics_2}
\end{figure}

\paragraph{Statistical Analysis on Domain-specific Performance Difference} The statistical results are presented in Figures \ref{fig:statistics_1} and \ref{fig:statistics_2}. Each row in either figure consists of four distinct sub-figures. The two sub-figures on the left side illustrate the performance of the respective model for abductive reasoning, while the two on the right side demonstrate the deductive performance. In each pair of two sub-figures, the barplot shows the weighted F1-score for each category across difficulty levels calculated using equation (4), while the heatmap displays the results of the chi-square test \cite{bolboacua2011pearson} with each cell corresponding to the p-value of the test regarding any pairwise categories. The application of chi-square test in this regard aims to determine whether there is a significant association between two distributions. As shown in each heatmap, the darker blue ($p-value = 0.05$ at different thresholds) implies a significant difference between the distributions of two categories, while the lightest blue ($p-value > 0.05$) suggests no significant association. 

Based on the barplots and heatmaps in Figures \ref{fig:statistics_1} and \ref{fig:statistics_2}, there are several observations to highlight in terms of models' performance. \textit{First}, GPT-4o, the series of Yi models, along with Qwen-7 and Qwen-14, exhibit a relatively higher weighted F1-score for these categories. In particular, GPT-4o and the series of Yi models demonstrate a higher weighted F1-score compared to the other models for abductive reasoning tasks. \textit{Second}, most of these models perform better in deductive reasoning tasks than abductive reasoning tasks. This observed pattern is consistent across most of the models and categories under investigation in this study. \textit{Third}, the models' performance varies significantly across different categories. For instance, based on the results of Yi-34 and Qwen-32, the weighted F1-score for the abstract category is much higher than that of other categories. However, it is also noted that the math category consistently displays a comparatively lower weighted F1-score across these categories. \textit{Fourth}, the abstract category is more likely to display significant differences when compared to other models, as demonstrated in the cases of Qwen-0.5, Qwen-1.5, Qwen-14, Qwen-32, Qwen-110, Yi-9, Yi-34, and GPT-4o.

It is possible that some may question whether the observed differences in model performance are due to the varying input lengths, rather than the effect of different instantiations. To address this potential concern, we have conducted a series of experiments to investigate the correlation between input length and model performance. The results of these experiments, which can be found in Appendix \ref{app:length_correlation}, suggest that there is little to no correlation between the two.

\subsection{Generalization by Finetuning}
We use finetuning to study the generalization ability from data. The following research questions are addressed: (1) How does the generalization potential of abstract data compare with contextualized data? Specifically, can a model trained on abstract logic instances generalize to contextualized logical reasoning text, and vice versa? (2) What is the impact of model scaling on generalization performance, and does this impact vary depending on the type of data used for fine-tuning (abstract vs. contextualized)? (3) Can models trained on instantiated data in one domain generalize to other domains, and does the diversity of domains in the fine-tuning data impact generalization performance?

\begin{figure}[!ht]
    \centering
    \includegraphics[scale=0.27]{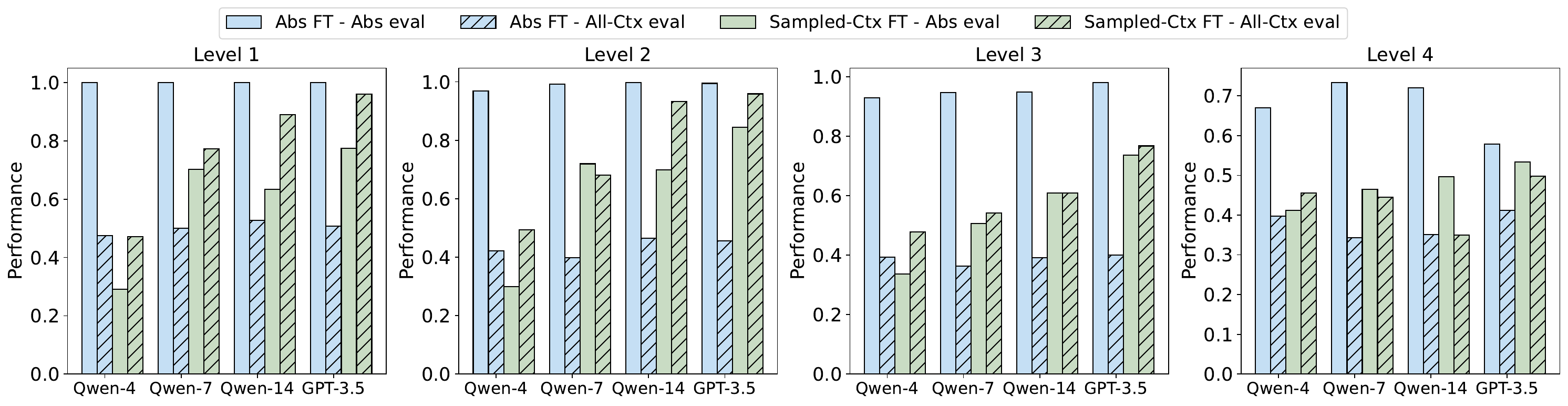}
    \caption{Performance Generalization Using Abstract Data and Sample-ctx Data.}
    \label{fig:abs_vs_ctx}
\end{figure}

\paragraph{Abstract vs. Contextualized}
To compare the generalization ability of abstract and contextualized data, four models (Qwen-4, Qwen-7, Qwen-14, and GPT-3.5-turbo) are fine-tuned on both purely abstract and purely contextualized data. The benchmark consists of 256 formal logic templates, resulting in 1280 abstract data points. While there are significantly more contextualized data points (1280 * 11), a random sample of 1280 is used for fine-tuning, with each logic template having approximately five instantiations. This subset of contextualized data is referred to as sampled-ctx to distinguish it from the whole contextually instantiated datapoints.

The results indicate that when fine-tuning on abstract data alone, although it learns well on abstract logic samples, it struggles to generalize to contextualized data, with performance decreasing as the dataset difficulty level increases. Conversely, fine-tuning on sampled-ctx data leads to significant improvements in general performance. The performance on purely abstract data is greatly improved. Furthermore, substantial performance improvements are observed in all contextualized data, with near-perfect weighted F1 results for contextualized data in level 1 and 2 in GPT-3.5. It should be noted that while some data points in the evaluation set are present in the fine-tuning set (sampled-ctx), they constitute only a small fraction (1/11) of the contextualized data in the evaluation set.

However, the results of the fine-tuning experiments for the most challenging level (level 4) were found to be unsatisfactory. The performance of the Qwen-14 model on contextualized data is observed to be even slightly worse than that of the same model fine-tuned on purely abstract data. This finding suggests that it is significantly more difficult to uncover the underlying logic reasoning pattern in highly complex reasoning tasks using contextualized data, even when fine-tuning on multiple such instantiations. While contextualized-data fine-tuning may be beneficial for simpler levels, it may be more straightforward to utilize purely abstract logic data to discern the underlying logic pattern in cases of extreme complexity.

\paragraph{Model Scale Effect on Generalization}
The impact of model scaling on generalization performance varies depending on the type of data used for fine-tuning. When fine-tuning on abstract data, larger models provide only marginal improvements in performance, suggesting that the potential of abstract data for generalization is limited. However, when fine-tuning on sampled-ctx data, the rate of improvement in evaluation performance increases in relation to model size noticeably when evaluated on both abstract and all-contextualized data, indicating the potential of contextualized data for learning underlying patterns and generalization.

\begin{wrapfigure}{R}{0.43\textwidth}
\centering
\includegraphics[scale=0.43]{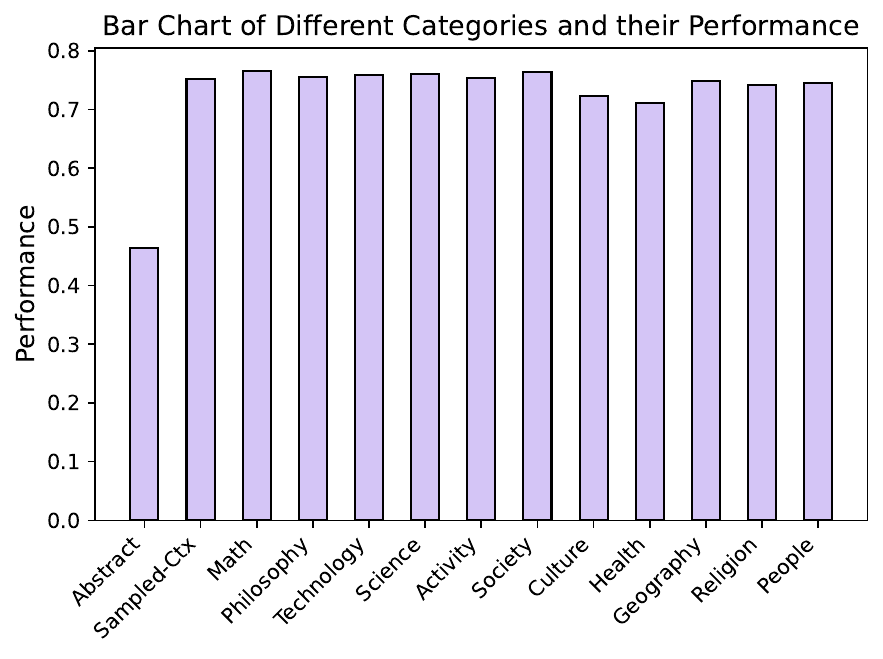}
\caption{Model performance by finetuning on different domains.}
\end{wrapfigure}

\paragraph{Single-domain vs. Multi-domain}

One potential explanation for the superior generalization capacity of sampled-ctx data is the diversity of domains from which it is sampled, as opposed to the relative homogeneity of abstract data. To investigate the impact of domain diversity, GPT-3.5 is fine-tuned on various single-domain instantiated datasets, each containing 1280 data points, and evaluated on the entire instantiated and abstract datasets.

The results show that the performance of models fine-tuned on different single-domain instantiated datasets is similar, and in some cases, even better than that of models fine-tuned on sampled-ctx data. This suggests that domain diversity does not play a significant role in the observed generalization capacity.


\section{Conclusion}
This paper presents a thorough investigation into the logic reasoning abilities of large language models. By utilizing the \method benchmark, we were able to disentangle logic reasoning from text understanding when performing reasoning tasks. Our results demonstrate that a model's performance on a given reasoning task can be significantly influenced by the context or domain in which the variables of the task are situated.

Furthermore, we examined the data generalization capabilities of abstract and instantiated data. Our findings reveal an intriguing pattern: instantiated data exhibits a greater potential for logic reasoning generalization in fine-tuning, regardless of the domain or number of domains utilized. This suggests that instantiated data may be more effective for fine-tuning models to perform well on a wide range of reasoning tasks.

In summary, our study sheds light on the factors that impact the logic reasoning abilities of large language models and provides insights into how to improve their performance. Future work could explore the use of instantiated data for fine-tuning models on more complex reasoning tasks, as well as the development of benchmarks that better capture the nuances of real-world reasoning scenarios.

\section{Appendix}

\subsection{Data Examples}
\label{app:data_examples}
The following table presents several examples showing abductive and deductive reasoning with their respective difficulty levels and domains. The left column shows examples of abstract instantiations, while the right column shows contextually instantiated examples in specific domains.

\begin{center}
\begin{longtable}{p{0.25\textwidth} p{0.68\textwidth}}
\caption[Examples of abductive and deductive reasoning.]{Examples of abductive and deductive reasoning.} \label{grid_mlmmh} \\

\hline \multicolumn{1}{c}{\textbf{Abstract Example}} & \multicolumn{1}{c}{\textbf{Specific Domain Example}} \\
\endfirsthead

\multicolumn{2}{c}%
{{\bfseries \tablename\ \thetable{} -- continued from previous page}} \\
\hline \multicolumn{1}{c}{\textbf{Abstract Example}} &
\multicolumn{1}{c}{\textbf{Specific Domain Example}} \\ \hline 
\endhead

\hline \multicolumn{2}{r}{{Continued}} \\ \hline
\endfoot

\hline \hline
\endlastfoot

\midrule
\multicolumn{2}{c}{Abductive Reasoning} \\
\midrule
Level 1 - Abstract &  Level 1 - Geography and Places \\
\midrule
aaa: vxkgr  & aaa: The terrain has experienced significant uplift.  \\
aab: caunc  & aab: Powerful erosional forces have shaped the land.  \\
aac: ybyz  & aac: The area features tall, steep mountains.  \\
reasoning task: (vxkgr or caunc) - ybyz. Given ybyz is False, what is the value of caunc? & reasoning task: If an area of land has experienced significant uplift or been shaped by powerful erosional forces, then the terrain will feature tall, steep mountains. Given that the area does not have tall, steep mountains, can it be determined if powerful erosional forces have shaped the land?  \\

\midrule
Level 2 - Abstract & Level 2 - Mathematics and Logic \\
\midrule
aaa: ttjmx & aaa: The prior probability is a uniform distribution \\
aab: kottz & aab: The prior probability expresses existing beliefs about the parameters. \\
aac: wqeq & aac: A prior probability distribution is specified. \\
aad: mnze & aad: New data is collected. \\
aae: zkx & aae: The posterior probability is calculated using Bayes' theorem. \\
aaf: pofk & aaf: The posterior probability provides an improved estimate of the parameters. \\
reasoning task: (wqeq or mnze) - zkx. (NOT ttjmx) - kottz. (kottz or zkx) - pofk. Given pofk is False, what is the value of ttjmx? & reasoning task: In Bayesian statistics, if a prior probability distribution is specified or new data is collected, then the posterior probability can be calculated using Bayes' theorem to update the probability based on the new evidence. If the prior probability is not a uniform distribution, then it expresses existing beliefs or knowledge about the values of the parameters. If the prior probability expresses existing beliefs or the posterior probability is calculated, then there is sufficient information to update the probability distribution. Given that the statement "The posterior probability provides an improved estimate of the parameters" is false, can it be determined whether the prior probability is a uniform distribution or not? \\

\midrule
Level 3 - Abstract & Level 3 - Technology and Applied Sciences \\
\midrule
aaa: dmacf & aaa: Regular vulnerability scans are performed. \\
aab: my & aab: Penetration testing is conducted quarterly. \\
aac: qnvj & aac: Security weaknesses are proactively identified. \\
aad: lxnf & aad: Operating systems are up to date with patches. \\
aae: jf & aae: Antivirus software is installed on all computers. \\
aaf: ors & aaf: Endpoint devices are protected. \\
aag: kuyl & aag: The overall attack surface is minimized. \\
aah: jal & aah: The firewall is properly configured. \\
aai: rqo & aai: Intrusion detection systems are active. \\
aaj: vrmxo & aaj: The network perimeter is secure. \\
aak: mcwe & aak: Employees have completed security training. \\
aan: pdzyf & aan: Security policies are strictly enforced. \\
aao: guwls & aao: Employees follow secure computing practices. \\
aap: xjgwm & aap: Internal systems and data are well-defended. \\
aaq: vv & aaq: The organization has strong cybersecurity posture. \\
reasoning task: (wqeq or mnze) - zkx. (NOT ttjmx) - kottz. (kottz or zkx) - pofk. Given pofk is False, what is the value of ttjmx? & reasoning task: If the firewall is properly configured or intrusion detection systems are active, then the network perimeter is secure. When employees have completed security training and security policies are strictly enforced, it implies that employees follow secure computing practices. If the network perimeter is secure or employees follow secure practices, then internal systems and data are well-defended. Having up-to-date operating systems with the latest patches or antivirus software installed on all computers means the endpoint devices are protected. Performing regular vulnerability scans or conducting quarterly penetration testing allows security weaknesses to be proactively identified. If security weaknesses are proactively identified or endpoint devices are protected, then the overall attack surface is minimized. When the attack surface is minimized and internal systems and data are well-defended, it indicates the organization has a strong cybersecurity posture. Given that the organization does not have a strong cybersecurity posture, can it be determined if operating systems are up to date with patches? \\

\midrule
Level 4 - Abstract & Level 4 - Culture and Arts \\
\midrule
aaa: cg & aaa: Sophie cannot practice her beam routine. \\
aab: ysjeo & aab: Sophie needs to prepare new skills. \\
aac: uby & aac: Sophie requires dedicated practice time. \\
aad: vwwf & aad: The springboard is broken. \\
aae: lj & aae: The vault is not stable. \\
aaf: qd & aaf: Performing vault runs is risky. \\
aag: miz & aag: Sophie is not able to practice effectively. \\
aah: tfxbc & aah: Sophie's coach is at practice. \\
aai: aaw & aai: Sophie does not have supervision. \\
aaj: oftr & aaj: Sophie is allowed to train. \\
aak: fzsq & aak: Sophie is not making progress in her gymnastics. \\
aan: yxt & aan: The balance beam is set up properly. \\
aao: ln & aao: Sophie cannot practice her beam routine. \\
aap: qa & aap: The uneven bars are not at the correct height. \\
aaq: py & aaq: Sophie cannot work on her bar skills. \\
aar: qe & aar: Sophie faces a major hindrance to her practice. \\
aas: ng & aas: The floor mat has tears and needs to be replaced. \\
aat: bhjb & aat: The floor area is not large enough for a full floor routine. \\
aau: djay & aau: It is unsafe for Sophie to practice floor exercises. \\
aav: pvize & aav: With an upcoming competition, Sophie needs to prepare new skills. \\
aaw: tk & aaw: Sophie does not have enough energy to train effectively. \\
aax: vod & aax: Sophie's gymnastics career is at risk.\\
aay: dngja & aay: Sophie's gymnastics performance will likely be impacted negatively. \\
aaz: ozyue & aaz: Sophie may need to consider withdrawing from competitions. \\
reasoning task: (NOT yxt) -> ln. (NOT qa) -> py. (ln or py) -> qe. (ng or bhjb) -> djay. (vwwf or lj) -> qd. (NOT tfxbc) -> aww. (NOT aww) -> oftr. (NOT pvize) -> tk. (cg or ysjeo) -> uby. (uby or qd) -> miz. (miz or oftr) -> fzsq. (djay or tk) -> vod. (qe or vod) -> dngja. (fzsq or dngja) -> ozyue. Given ozyue is False, what is the value of yxt? & reasoning task: The balance beam not being set up properly means Sophie cannot practice her beam routine. Similarly, if the uneven bars are not at the correct height, Sophie cannot work on her bar skills. If Sophie is unable to train on at least one apparatus, she faces a major hindrance to her practice. Torn floor mats needing replacement or insufficient floor space makes it unsafe for Sophie to practice floor exercises. A broken springboard or unstable vault makes performing vault runs risky. If it is unsafe to practice floor or vault exercises, Sophie cannot train safely or productively. Sophie's coach not being at practice means she does not have supervision. Having supervision allows Sophie to train. If Sophie did not fuel properly before practice, she will not have enough energy to train effectively.  With an upcoming competition, Sophie needs to prepare new skills, requiring dedicated practice time. If Sophie's training is compromised by risky apparatus or lack of practice time, she will not be able to practice effectively. If Sophie's training session is unproductive or she faces major hindrances, then she is not making progress in her gymnastics. Lack of progress or likely negative performance impacts put Sophie's gymnastics career at risk. Given that Sophie is not considering withdrawing from competitions, what can be determined about the balance beam being set up properly?\\

\midrule
\multicolumn{2}{c}{Deductive Reasoning} \\
\midrule
Level 1 - Abstract & Level 1 - Natural and Physical Sciences \\
\midrule
aaa: pusvu & aaa: A cold front is approaching the region \\
aab: hs & aab: A warm air mass is stagnant over the area \\
aac: ivl & aac: Atmospheric instability is likely to develop \\
reasoning task: pusvu is True. hs is False. (pusvu or hs) - ivl. Deduce the result of ivl. & reasoning task: A cold front is approaching the region, but there is no warm air mass stagnant over the area. If a cold front approaches or a warm air mass is stagnant, then atmospheric instability is likely to develop. Can we say that atmospheric instability will likely develop in this scenario? \\

\midrule
Level 2 - Abstract & Level 2 - Society and Social Sciences \\
\midrule
aaa: jd & aaa: John Lee was born in the United States \\
aab: bfk & aab: John Lee's parents immigrated from South Korea \\
aac: wng & aac: John Lee has Korean ancestry \\
aad: vko & aad: The Lee family speaks Korean fluently \\
aae: cva & aae: The Lee family identifies as Korean-American \\
aaf: qymwa & aaf: The Lee family has a connection to Korean culture \\
aag: cr & aag: John Lee is considered Korean-American \\
reasoning task: cva is True. vko is False. (vko or cva) - qymwa. jd is True. bfk is True. (jd or bfk) - wng. (wng and qymwa) - cr. Deduce the result of cr. & reasoning task: The Lee family identifies as Korean-American, but they do not speak Korean fluently. If the Lee family speaks Korean fluently or identifies as Korean-American, then they have a connection to Korean culture. John Lee was born in the United States, and his parents immigrated from South Korea. If John Lee was born in the U.S. or his parents immigrated from South Korea, then he has Korean ancestry. If John Lee has Korean ancestry and his family has a connection to Korean culture, then he is considered Korean-American. Can we conclude that John Lee is considered Korean-American based on the given information? \\

\midrule
Level 3 - Abstract & Level 3 - Culture and Arts \\
\midrule
aaa: rfx & aaa: The opera house was empty \\
aab: gurl & aab: The soprano sang the aria beautifully \\
aac: imnsi & aac: Some people attended the opera \\
aad: wjgx & aad: The sets malfunctioned \\
aae: tg & aae: The costumes were delivered late \\
aaf: kopg & aaf: There were technical difficulties \\
aag: khh & aag: The show faced some challenges \\
aah: ozro & aah: The orchestra played flawlessly \\
aai: pg & aai: The tenor forgot his lines \\
aaj: bill & aaj: The performance went smoothly \\
aak: mek & aak: There was a major disruption \\
aan: jp & aan: The opening night was eventful \\
reasoning task: gurl is True. pg is False. ozro is True. (ozro or pg) - bill. rfx is False. (rfx or gurl) - imnsi. tg is False. wjgx is False. (wjgx or tg) - kopg. (imnsi or kopg) - khh. (NOT bill) - mek. (khh or mek) - jp. Deduce the result of jp. & reasoning task: The soprano sang her aria beautifully and the orchestra played flawlessly, but the tenor forgot his lines. If the orchestra played well or the tenor forgot his lines, then the performance did not go entirely smoothly. The opera house was not empty since the soprano's beautiful aria meant some people attended. The costumes were not delivered late and the sets did not malfunction, so there were no technical difficulties. If some people attended or there were technical difficulties, the show would have faced some challenges. Since the performance did not go smoothly, it implies there was a major disruption. If the show faced challenges or had a major disruption, the opening night of this opera was quite eventful. Given this, was the opening night of the opera eventful? \\

\midrule
Level 4 - Abstract & Level 4 - Health and Fitness \\
\midrule
aaa: msta & aaa: Sue did push-ups yesterday \\
aab: fo & aab: Sue did not do pull-ups yesterday \\
aac: jfnrh & aac: Sue did some upper body exercises yesterday \\
aad: ssb & aad: Sue did squats yesterday \\
aae: ac & aae: Sue did not do squats yesterday \\
aaf: dzda & aaf: Sue only trained upper body yesterday \\
aag: hujcf & aag: Sue did burpees yesterday \\
aah: pil & aah: Sue did not do burpees yesterday \\
aai: dyue & aai: Sue trained her core muscles yesterday \\
aaj: sgniu & aaj: Sue did planks yesterday \\
aak: stbf & aak: Sue had an effective core workout yesterday \\
aan: pswg & aan: Sue did an intense workout yesterday \\
aao: fkyxi & aao: Sue had a focused or intense workout yesterday \\
aap: outm & aap: Sue did lunges yesterday \\
aaq: ybjj & aaq: Sue did step-ups yesterday \\
aar: eek & aar: Sue trained her leg muscles yesterday \\
aas: wmejd & aas: Sue did wall sits yesterday \\
aat: rdbk & aat: Sue did not do calf raises yesterday \\
aau: rqmc & aau: Sue did some quad and hamstring exercises yesterday \\
aav: bw & aav: Sue had an effective lower body workout yesterda \\
aaw: xvd & aaw: Sue did not do a full body workout yesterday \\
aax: pg & aax: Sue did a partial body workout yesterday \\
aay: qbli & aay: Sue did a full body workout yesterday \\
aaz: qvb & aaz: Sue had a comprehensive workout yesterday \\
reasoning task: fo is False. msta is True. (msta or fo) -> jfnrh. dyue is True. xvd is False.  (NOT xvd) -> pg. ssb is True. (NOT ssb) -> ac. (jfnrh and ac) -> dzda. sgniu is True. (dyue or sgniu) -> stbf. outm is True.   rdbk is False. ybjj is True. (outm or ybjj) -> eek. wmejd is True. (wmejd or rdbk) -> rqmc. (eek and rqmc) -> bw. (NOT pg) -> qbli. (bw or qbli) -> qvb. hujcf is True. (NOT hujcf) -> pil. (pil and stbf) -> pswg. (dzda or pswg) -> fkyxi. (fkyxi or qvb) -> abc. Deduce the result of abc. & reasoning task: Sue did push-ups but not pull-ups yesterday. If she did push-ups or pull-ups, then she did some upper body exercises. Sue trained her core by doing planks. Since she did not do a full body workout, it means she did a partial body workout. Sue did squats yesterday, so it is not true that she did not do squats.  If Sue did some upper body exercises and did not do squats, then she only trained upper body. If Sue did planks or trained her core muscles, then she had an effective core workout. Sue did lunges and step-ups, but not calf raises. If she did lunges or step-ups, then she trained her leg muscles. If Sue did wall sits or calf raises, then she did some quad and hamstring exercises. If Sue trained her leg muscles and did some quad/hamstring exercises, then she had an effective lower body workout. If Sue did not do a partial body workout, then she did a full body workout.If Sue had an effective lower body workout or did a full body workout, then she had a comprehensive workout. Sue did burpees yesterday, so it is not true that she did not do burpees. If Sue did not do burpees and had an effective core workout, then she did an intense workout. If Sue only trained upper body or did an intense workout, then she had a focused or intense workout. If Sue had a focused/intense workout or a comprehensive workout, then she had a productive bodyweight training session. Did Sue have a productive bodyweight training session yesterday?\\

\end{longtable}
\end{center}

\subsection{Length Correlation}
\label{app:length_correlation}
Some maybe curious whether the performance degradation and variance on instantiated data are correlated with input text length. To see whether there is indeed a correlation between model performance and text length, we employ four models of varying sizes (Qwen-0.5, Qwen-7, Qwen-32, Qwen-110) and conduct length-based performance ablation. We then analyze the performance of each model based on the length of the input text. Specifically, for each graph presented below, the x-axis represents the text length, while the y-axis represents the corresponding model performance. The y-axis value for each x-axis value $x$ is the model performance on the part of the data whose corresponding input text length is smaller than $x$. 

\begin{figure}[!ht]
    \centering
    \includegraphics[scale=0.35]{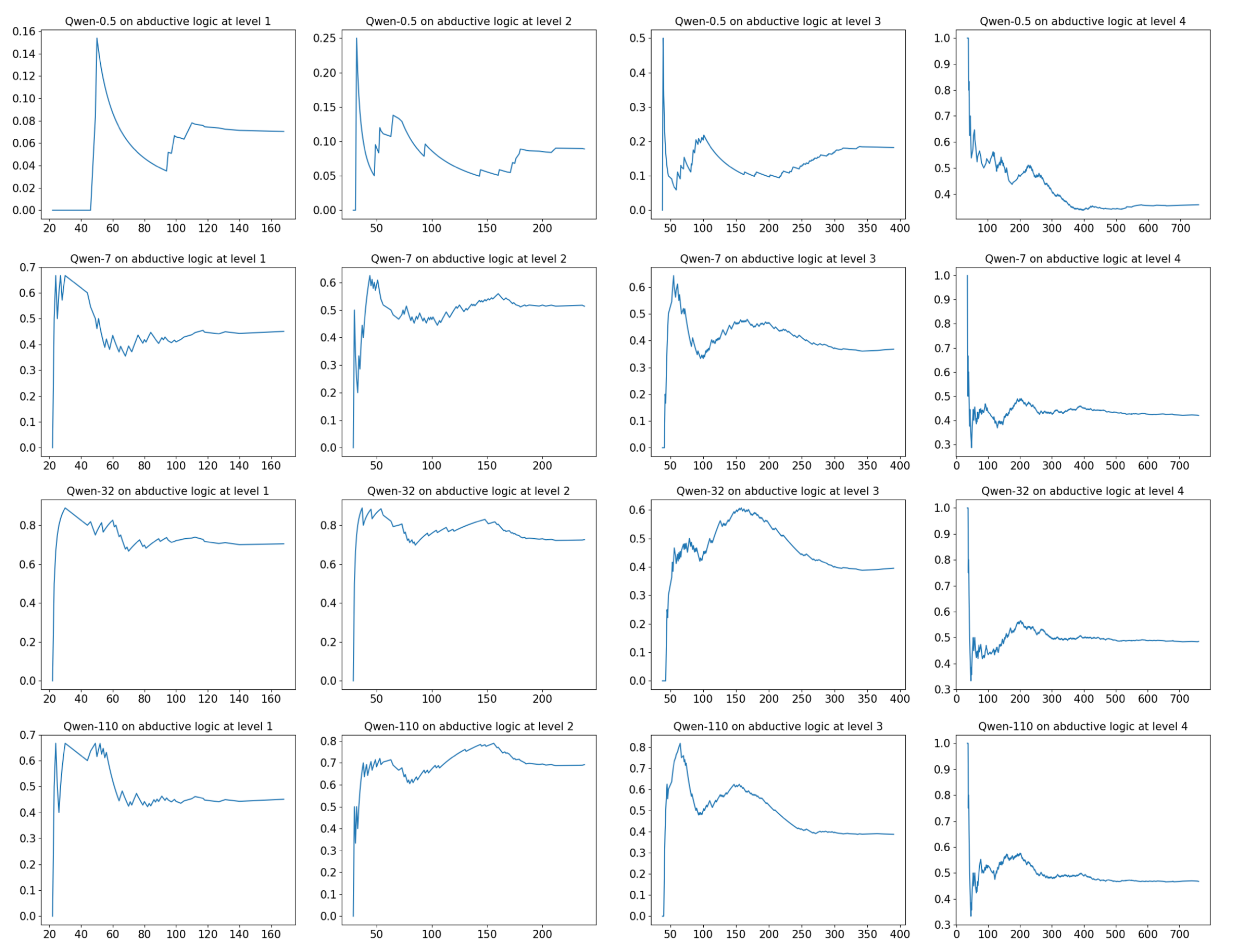}
    \caption{Length-based performance collection on abductive logic. The four rows correspond to four models, and four columns correspond to four difficulty levels.}
    \label{fig:abductive_length_correlation}
\end{figure}

\begin{figure}[!ht]
    \centering
    \includegraphics[scale=0.35]{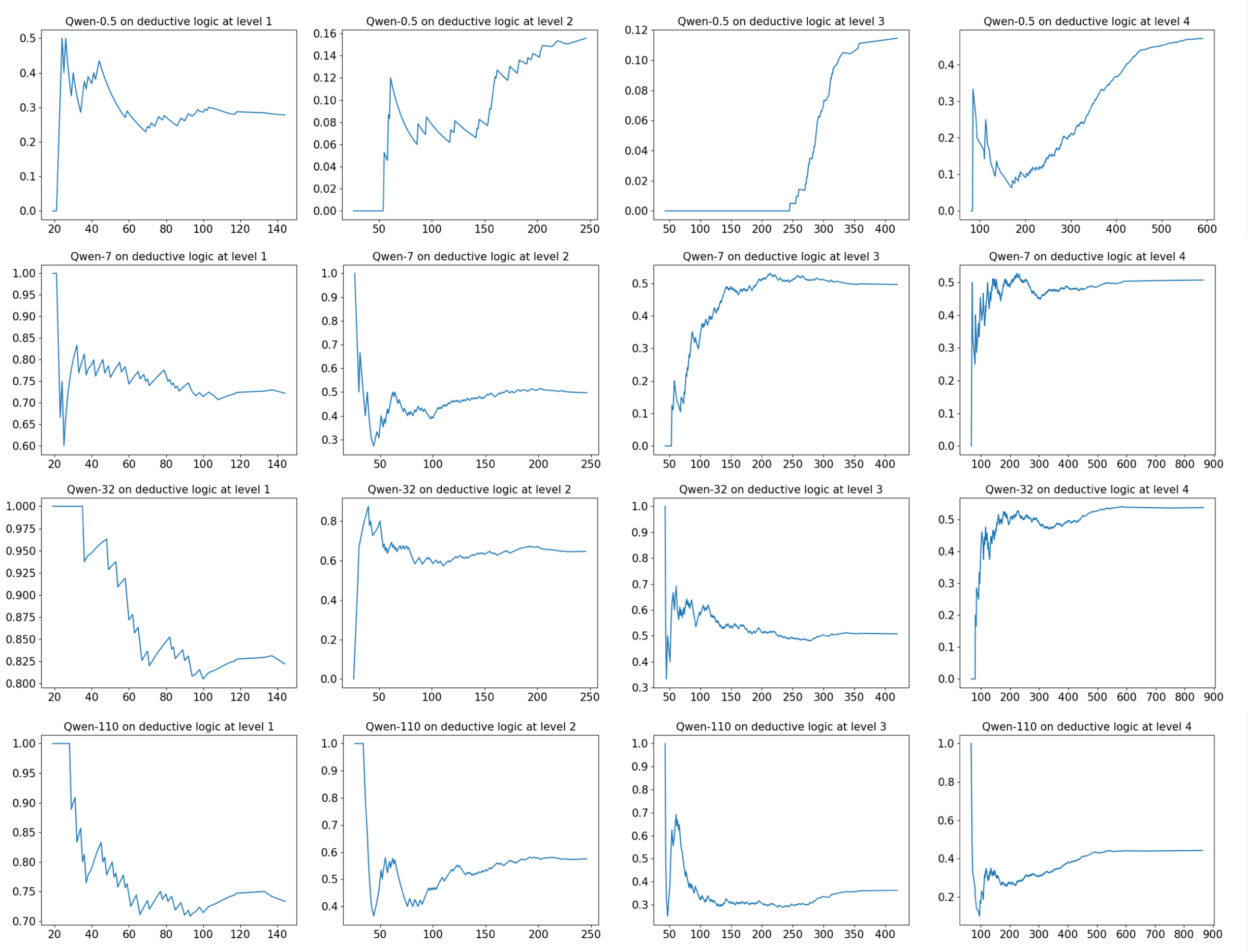}
    \caption{Length-based performance collection on deductive logic. The four rows correspond to four models, and four columns correspond to four difficulty levels.}
    \label{fig:deductive_length_correlation}
\end{figure}

Based on the two images Figure\ref{fig:abductive_length_correlation} and Figure \ref{fig:deductive_length_correlation}, we cannot see any consistent correlation between model performance and input length after tokenization using model corresponding tokenizer. 

\subsection{Human verification}
\label{app:verification}
The benchmark dataset used in this study was synthetically generated based on the Claude-3 model, but was subsequently reviewed by a panel of five annotators, each holding a Ph.D. degree in a diverse range of fields, including computer science, informatics, civil engineering, and medicine. Each annotator was responsible for reviewing 2-3 domains, with each domain consisting of 20 data datapoints for each level of the dataset. In total, 220 datapoints were manually reviewed for each level of the dataset, comprising 10\% of the total dataset.

The following table \ref{tab:human_verification} presents the rate of correctness for template adherence and fact reckoning, respectively, as determined by the panel of annotators.

\begin{table}[!ht]
    \centering
    \begin{tabular}{c|c|c}
    \toprule
       data level  & template adherence & fact reckoning \\
       \hline
        1 & 100\% & 95.45\%\\
        \hline
        2 & 100\% & 94.54\%\\
        \hline
        3 & 100\% & 96.36\%\\
        \hline
        4 & 96.36\% & 93.18\%\\
        \bottomrule
    \end{tabular}
    \caption{Accuracy of Synthetically-generated Data}
    \label{tab:human_verification}
\end{table}

The high rates of correctness for both template adherence and fact reckoning suggest that the synthetically generated dataset is of high quality and is suitable for use in evaluating the performance of LLMs on instantiated logic tasks.

\bibliographystyle{unsrt}  
\bibliography{references} 

\end{document}